\documentclass[journal]{IEEEtran}
\ifCLASSINFOpdf
\else
\fi

\usepackage{graphicx}
\usepackage{fixltx2e}
\usepackage[latin9]{inputenc}
\usepackage{float}
\usepackage{multirow}
\usepackage{algorithm}
\usepackage{algorithmic}
\usepackage{amsmath,amsthm,amssymb}
\usepackage{xcolor}
\usepackage{subfig,wrapfig}
\usepackage{rotating}
\usepackage{url}
\usepackage{booktabs}
\usepackage{rotating}
\usepackage{indentfirst}
\usepackage{caption}

\begin{document}
%
\title{A Good Practice Towards Top Performance of Face Recognition: Transferred Deep Feature Fusion}
%
%
%
\author{Lin~Xiong$^1$$^*$$^\dagger$, Jayashree~Karlekar$^1$$^*$, Jian~Zhao$^2$$^*$$^\dagger$ ~\IEEEmembership{Student Member,~IEEE}, Yi Cheng$^1$, Yan Xu$^1$, Jiashi~Feng$^2$ ~\IEEEmembership{Member,~IEEE}, Sugiri~ Pranata$^1$, and Shengmei~ Shen$^1$

\thanks{$^1$L. Xiong, J. Karlekar, Y. Cheng, Y. Xu, S. Pranata and S.M. Shen are with Panasonic R\&D Center Singapore, Singapore \protect(lin.xiong, karlekar.jayashree, yi.cheng, yan.xu, sugiri.pranata, shengmei.shen)@sg.panasonic.com.}
\thanks{$^2$J. Zhao and J.S. Feng are with Department of Electrical and Computer Engineering, National University of Singapore, Singapore \protect (zhaojian90@u.nus.edu; elefjia@nus.edu.sg). J. Zhao was an intern at Panasonic R\&D Center Singapore during this work.}
\thanks{$^*$ L. Xiong, J. Zhao and J. Karlekar make an equal contribution.}
\thanks{$^\dagger$ L. Xiong and J. Zhao are the corresponding author.}

}
\markboth{IEEE Transactions on XXXX,~Vol.~XX, No.~XX, XX~201X}
{XIONG \MakeLowercase{\textit{et al.}}: A GOOD PRACTICE TOWARDS TOP PERFORMANCE OF FACE  RECOGNITION: TRANSFERRED DEEP FEATURE FUSION}
\maketitle
\begin{abstract}
Unconstrained face recognition performance evaluations have traditionally focused on Labeled Faces in the Wild (LFW) dataset for imagery and the YouTubeFaces (YTF) dataset for videos in the last couple of years. Spectacular progress in this field has resulted in saturation on verification and identification accuracies for those benchmark datasets. In this paper, we propose a unified learning framework named Transferred Deep Feature Fusion (TDFF)  targeting at the new IARPA Janus Benchmark A (IJB-A) face recognition dataset released by NIST face challenge. The IJB-A dataset includes real-world unconstrained faces from 500 subjects with full pose and illumination variations which are much harder than the LFW and YTF datasets.
Inspired by transfer learning, we train two advanced deep convolutional neural networks (DCNN) with two different large datasets in source domain, respectively. By exploring the complementarity of two distinct DCNNs, deep feature fusion is utilized after feature extraction in target domain. Then, template specific linear SVMs is adopted to enhance the discrimination of framework. Finally, multiple matching scores corresponding different templates are merged as the final results. This simple unified framework exhibits excellent performance on IJB-A dataset. Based on the proposed approach, we have submitted our IJB-A results to National Institute of Standards and Technology (NIST) for official evaluation. Moreover, by introducing new data and advanced neural architecture, our method outperforms the state-of-the-art by a wide margin on IJB-A dataset. 
\end{abstract}
\IEEEpeerreviewmaketitle

\section{Introduction}
\label{sec1:Introduction}
\IEEEPARstart{F}{ace} recognition performance using features of Deep Convolutional Neural Network (DCNN) have been dramatically improved in recent years. Many state-of-the-art algorithms claim very close \cite{taigman2014deepface},\cite{sun2015deeply} or even have surpassed \cite{sun2015deepid3}, \cite{schroff2015facenet},\cite{chen2016unconstrained} human performance on Labeled Faces in the Wild (LFW) dataset. The saturation in recognition accuracy for current benchmark dataset has come. In order to push the development of frontier in regarding to unconstrained face recognition, a new face dataset template-based IJB-A is introduced recently \cite{klare2015pushing}, whose setting and solutions are aligned better with the requirements of real applications. 

\begin{figure}
	\label{Figure1} 
	\centering 
	\vspace{-3mm}
	\subfloat[Face recognition over single image.]{
	    \includegraphics[scale=0.4]{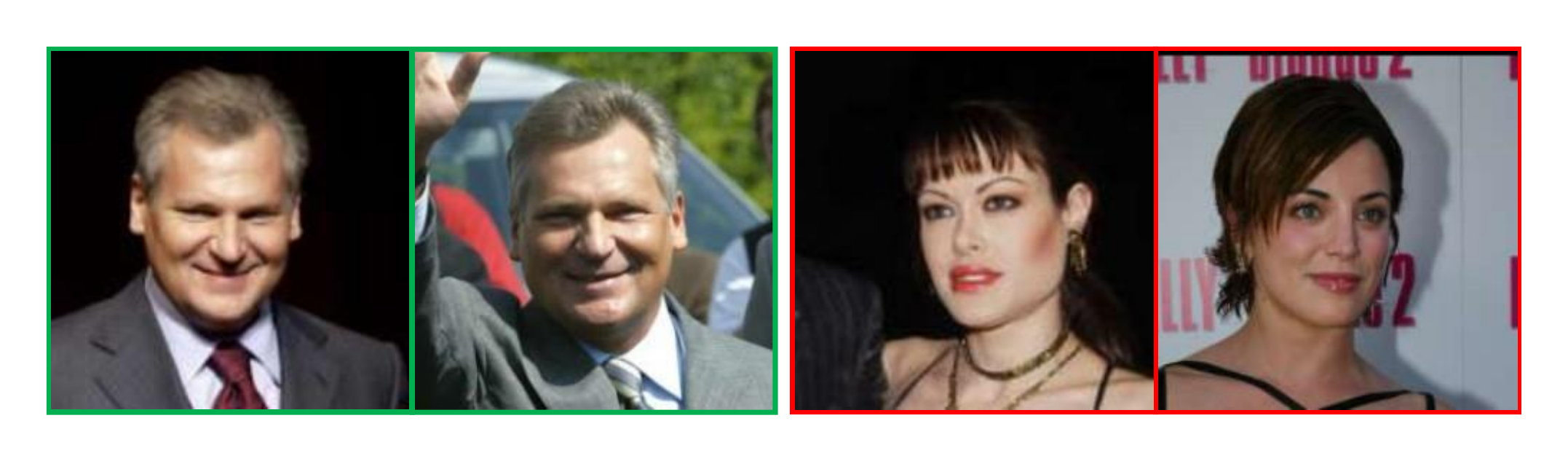} }
		\vspace{-5mm} 
		\label{fig:standard face recognition} 
	\subfloat[Unconstrained set-based face recognition.]{
	    \includegraphics[scale=0.4]{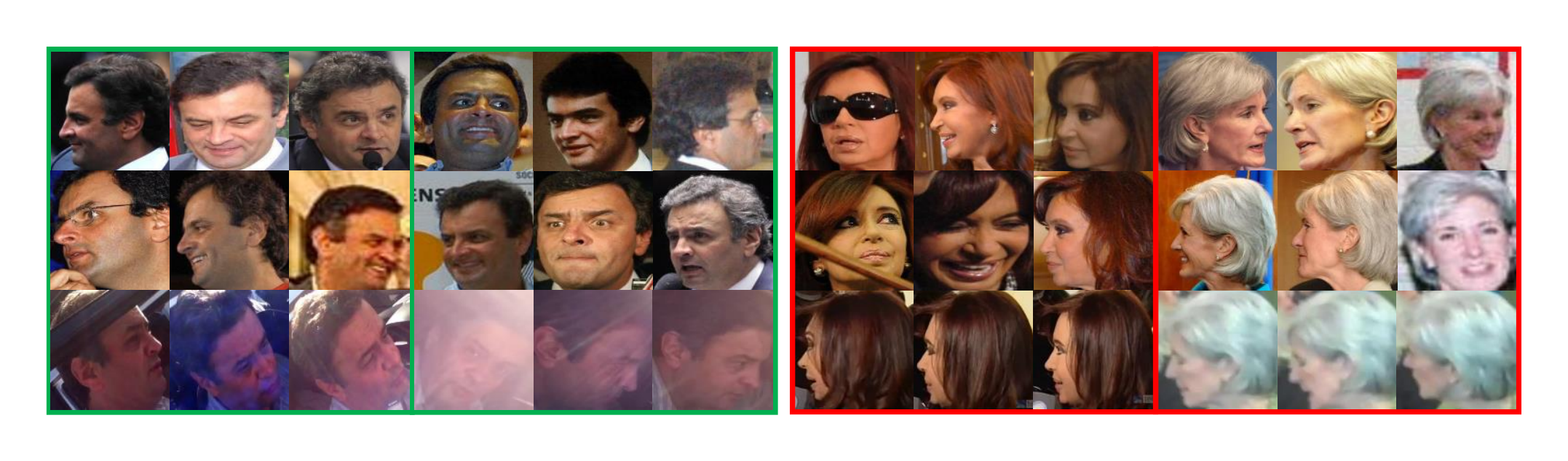}} 
		\label{fig:unconstrained set-based face recognition} 
	\small 
	\caption{\small{Comparison between face recognition over single image and unconstrained set-based face recognition. (a) Face recognition over single image. (b) Unconstrained set-based face recognition where each subject is represented by a set of mixed images and videos captured under unconstrained conditions. Each set contains large variations in face pose, expression, illumination and occlusion issues. Existing single-medium based recognition approaches cannot successfully address this problem consistently. Matched cases are bounded with green boxes, while non-matched cases are bounded with red boxes. Best viewed in color.}}\label{fig: Figure1} 
	\vspace{-0.1in} 
\end{figure}

The IJB-A dataset is created to provide the latest and most challenging dataset for both verification and identification as shown is Fig.1. Unlike LFW and YTF, this dataset includes both image and video of subjects manually annotated with facial bounding boxes to avoid the near frontal condition, along with protocols for evaluation of both verification and identification. Those protocols significantly deviate from standard protocols for many face recognition algorithms \cite{ye2016face},\cite{li2016robust}. Moreover, the concept of template is introduced, simultaneously. A template refers to a collection of all media (images and/or video frames) of an interested face captured under different conditions that can be utilized as a combined single representation for matching task. The template-based setting reflects many real-world biometric scenarios, where capturing a subject's facial appearance is possible more than once under different acquisition ways. In other words, this new IJB-A face recognition task requires to deal with a more challenging set-to-set matching problem successfully regardless of face capture settings (illumination, sensor, resolution) or subject conditions (facial pose, expression, occlusion).

 \begin{figure*}[t]
 	\label{Figure2}
 	\centering
 	\includegraphics[scale=0.7]{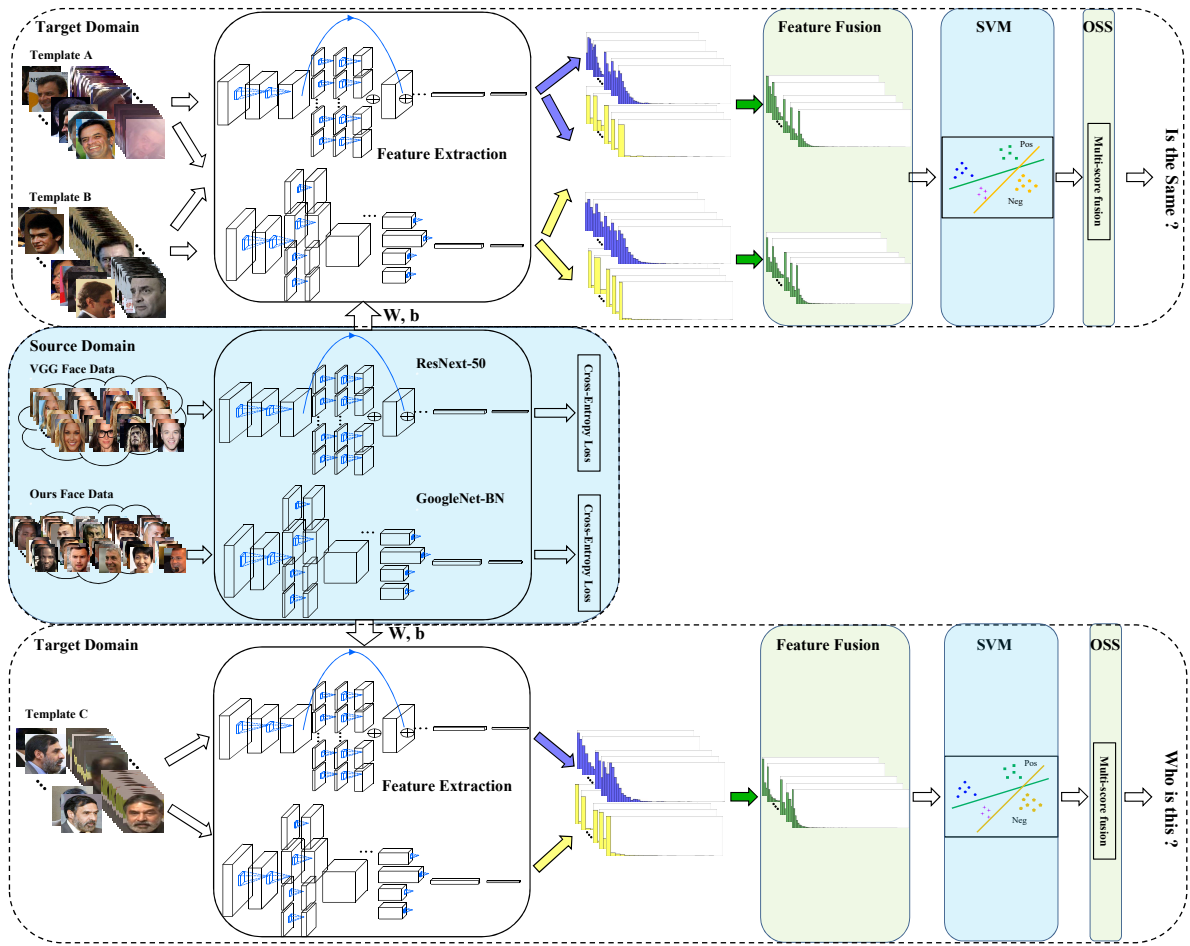}\\
 	\vskip -0.3in
 	\caption{{\small Framework overview. Our TDFF learning framework consists three components: Deep feature learning module locates middle component, Template-based unconstrained face recognition is included in upper and lower components. Training procedures are illustrated with blue blocks, two-stage fusion is depicted in green blocks. Best viewed in color.}}
 \end{figure*}

Our contributions can be summarized as following aspects:
\begin{enumerate}
	\item A unified learning framework named transferred deep feature fusion is proposed for face verification and identification.
	\item Two latest DCNN models are trained in source domain with two different large datasets  in order to take full advantage of complementary between models and datasets.
	\item Two-stage fusion are designed, one for features and another for similarity scores.
	\item One-vs-rest template specific linear SVMs with chosen negative set is trained in target domain.
\end{enumerate}

In this paper, we propose a unified learning framework named transferred deep feature fusion. It can effectively integrate superiority of each module and outperform the state-of-the-art on IJB-A dataset. Inspired by transfer learning \cite{pan2010survey}, facial feature encoding model of subjects are trained offline in a source domain, and this feature encoding model is transferred to a specific target domain where limited available faces of new subjects can be encoded. Specifically, in order to capture the intrinsic discrimination of subjects and enhance the generalization capability of face recognition models, we deploy two advanced deep convolutional neural networks (DCNN) with distinct architectures to learn the representation of faces on two different large datasets (each one has no overlap with IJB-A dataset) in source domain. These two DCNN models provide distinct feature representations which can better characterize the data distribution from different perspectives. The complementary between two distinct models is beneficial for feature representation \cite{sainath2015convolutional}. Thus, representing a face from different perspectives could effectively decrease ambiguity among subjects and enhance the generalization performance of face recognition especially on extremely large number of subjects. After offline training procedure, those two DCNN models are transferred to target domain where templates of IJB-A dataset as inputs are performed feature extraction with shared weights and biases, respectively. Then, features from two DCNN models are combined in order to obtain more discriminative representation. Finally, template specific linear SVMs are trained on fused features for classification. Furthermore, for set-to-set matching problem, multiple matching scores are merged into a single one \cite{masi2016pose},\cite{hassner2016pooling},\cite{masi2016we} for each template pair as the final results. Comprehensive evaluations on IJB-A public dataset well demonstrate the significant superiority of the proposed learning framework. Based on the proposed approach, we have submitted our IJB-A results to NIST for official evaluation. Furthermore, by introducing new data and advanced neural architecture, our method outperforms the state-of-the-art by a wide margin on IJB-A dataset. 

This paper is organized as follows. We review the related work in Section \uppercase\expandafter{\romannumeral2}.
Section \uppercase\expandafter{\romannumeral3} shows the details of transferred deep feature fusion. In Section \uppercase\expandafter{\romannumeral4}, a comprehensive evaluation on IJB-A dataset is shown.
Finally, the conclusion remarks and the future work are presented in Section \uppercase\expandafter{\romannumeral5}.

\section{Related Work}
\label{sec2:Related Work}
Recently, all the top performing methods for face recognition on LFW and YTF are all based on DCNN architectures. Such as the VGG-Face model \cite{parkhi2015deep}, as a typical application of the VGG-16 convolutional network architecture \cite{simonyan2014very} trained on a reasonably and publicly large face dataset of 2.6M images of 2622 subjects, provides state-of-the-art performance. This dataset is called as VGG-Face data for convenience in the following section. FaceNet \cite{schroff2015facenet} utilizes the DCNN with inception module \cite{szegedy2015going} for unconstrained face recognition. This network is trained using a private huge dataset of over 200M images and 8M subjects. DeepFace \cite{taigman2014deepface} deploys a DCNN coupled with 3D alignment, where facial pose is normalized by warping facial landmarks to a canonical position prior to encoding face images. DeepID2+ \cite{sun2015deeply} and DeepID3 \cite{sun2015deepid3} extend the FaceNet model by including joint Bayesian metric learning \cite{chen2012bayesian} and multi-task learning. More better unconstrained face recognition performance is provided by them. Moreover, DeepFace is trained using a private dataset of 4.4M images and 4,030 subjects. DeepID2+ and DeepID3 are trained also using a private dataset of 202,595 images and 10,117 subjects with 25 networks and 50 networks, respectively. The idea of multiple model ensemble is involved. Moreover, many approaches use metric learning in the form of triplet loss similarity or joint Bayesian for the final loss to learn an optimal embedding for face recognition \cite{schroff2015facenet},\cite{parkhi2015deep},\cite{chen2016unconstrained}. Thus, a recent study \cite{hu2015face} concludes that multiple networks ensemble and metric learning are crucial for improvement on LFW.   
\begin{figure}[h]
	\label{Figure3}
	\vskip -0.2in
	\centering
	\includegraphics[scale=0.33]{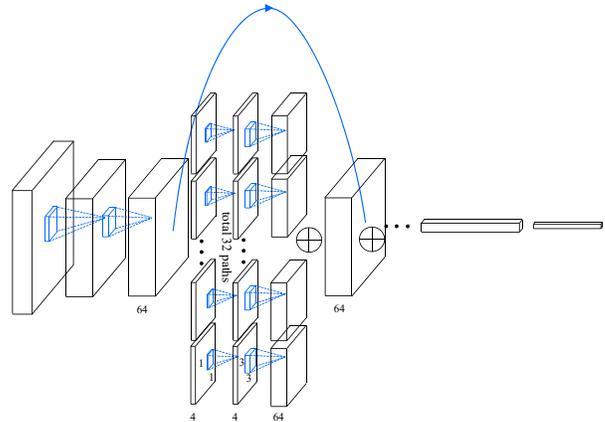}\\
	\caption{{\small A block of ResNeXt with cardinality=32.}}
	\vskip -0.1in
\end{figure}

\begin{figure}[h]
	\label{Figure4}
	\centering
	\includegraphics[scale=0.4]{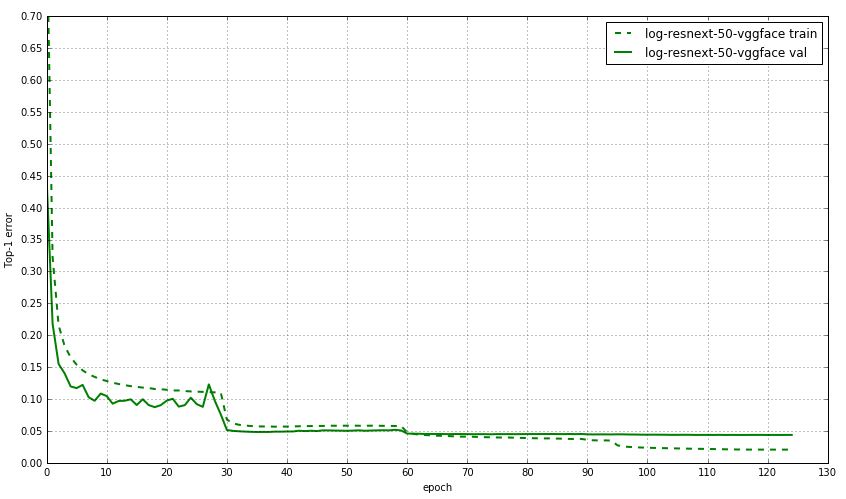}
	\caption{{\small Training on VGG-Face data. Solid curve denotes top 1 training error, and dotted line denotes validation error of the center crops.}}
	\vskip -0.1in
\end{figure}

\begin{figure}[h]
	\label{Figure5}
	\centering
	\includegraphics[scale=0.48]{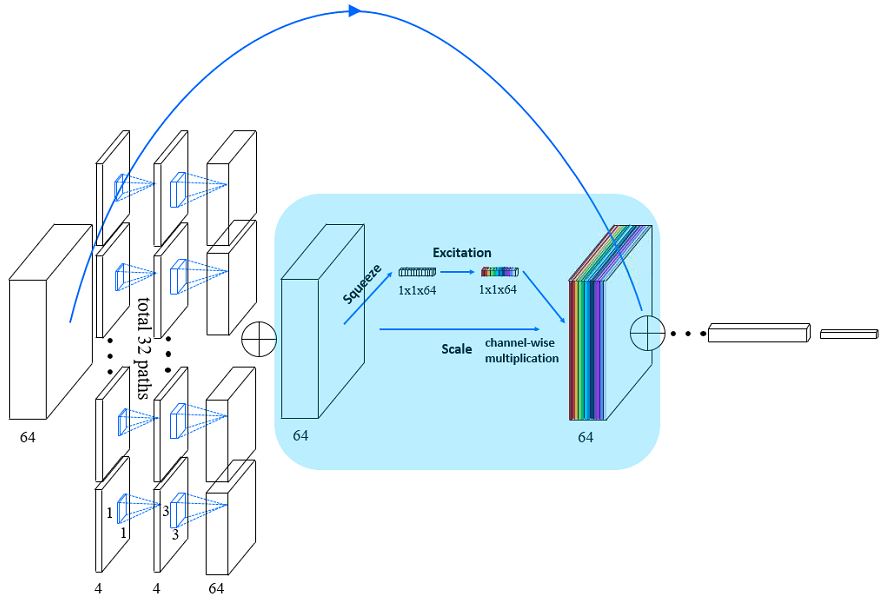}\\
	\caption{{\small A block of ResNeXt combined with Squeeze-and-Excitation, SE block is depicted in blue box. Best viewed in color.}}
	\vskip -0.1in
\end{figure}

With the advent of IJB-A dataset introduced by NIST in 2015, the task of template-based unconstrained face recognition has attracted extensive attention. So far as we known, most algorithms for this challenging problem are also based on DCNN architecture as top performing methods did on LFW and YTF. Chen \emph{et al.} \cite{chen2016unconstrained} achieve good performance by extracting feature representations via a DCNN trained on public dataset which includes 490,356 images and 10,548 subjects. And then, those features as inputs are applied to learn metric matrix in order to project the feature vector into a low-dimensional space, meanwhile, maximizing the between-class variation and minimizing within-class variation via joint Bayesian metric learning. B-CNN \cite{chowdhury2016one} applies the bilinear CNN architecture to face identification. Deep Multi-pose \cite{abdalmageed2016face} utilizes five pose specialized sub-networks with 3D pose rendering to encode multiple pose-specific features. Sensitivity of the recognition system to pose variations is reduced since an ensemble of pose-specific deep features is adopted. Pooling faces \cite{hassner2016pooling} aligns faces in 3D and bins them according to head pose and image quality. Pose-Aware Models (PAMs) \cite{masi2016pose} handles pose variability by learning Pose-Aware Models for frontal, half-profile and full-profile poses in order to improve face recognition performance in wild. 
Masi \emph{et al.} \cite{masi2016we} even question whether need to collect millions of faces or not for effective face recognition. Thus, a far more accessible means of increasing training data sizes is proposed. Pose, 3D shape and expression are utilized to synthesize more faces from CASIA-WebFace dataset \cite{yi2014learning}. Triplet Probabilistic Embedding (TPE) \cite{sankaranarayanan2016tripletP} couples a DCNN-based approach with a low-dimensional discriminative embedding learned using triplet probability constraints to solve the unconstrained face verification problem. TPE obtains better performance than previous algorithms on IJB-A dataset. Template Adaptation (TA) \cite{crosswhite2017template} proposes the idea of template adaptation which is a form of transfer learning to the set of media in a template. Combining DCNN features with template adaptation, it obtains better performance than TPE on IJB-A task. Ranjan \emph{et al.} propose an all-in-one method \cite{ranjan2017all} employed a multi-task learning framework that regularizes the shared parameters of CNN and builds a synergy among different domains and tasks. Until recently, Yang \emph{et al.} propose Neural Aggregation Network (NAN) \cite{yang2016neural} which produces a compact and fixed-dimension feature representation. It adaptively aggregates the features to form a single feature inside the convex hull spanned by them.  What's more interesting is that NAN learns to advocate high-quality face images while repelling low-quality ones such as blurred, occluded and improperly exposed faces. Thus, the face recognition performance on IJB-A dataset is pushed to reach an unprecedented height. Furthermore, Hayat \emph{et al.} proposes a joint registration and representation learning for unconstrained face identification \cite{hayat2017joint}, where the registration module based on spatial transformer network \cite{jaderberg2015spatial} and decision fusion are included. Moreover, Ranjan \emph{et al.} \cite{ranjan2017l2} add an 
$L_2$-constraint to the feature descriptors which restricts them to lie on a hypersphere of a fixed radius. Therefore, minimizing the softmax loss is equivalent to maximizing the cosine similarity for the positive pairs and minimizing it for the negative pairs. In this way, the verification performance on IJB-A dataset is refreshed again.

Last but not least, due to a simple yet powerful strategy to estimate target distribution and generate novel data is provided by the min-max two-player game \cite{goodfellow2014generative},\cite{denton2015deepgenerative}, many researches pay more and more attention to Generative Adversarial Network (GAN) from both the deep learning and computer vision domain. Especially, such as IJB-A task in unconstrained face recognition has very large facial pose variation, in other words, the facial pose distribution is usually unbalanced and has long-tail with extremely pose variations. By virtue of the idea of an adversarial loss for distribution modeling, the GAN can force the generated images to be, in principle, indistinguishable from real images. So, there are mainly two ways for alleviating the issue of facial pose unbalance. The one comes from \cite{zhao2017dual}, Dual-Agent Generative Adversarial Network (DA-GAN) can improve the realism of a face simulator's output using unlabeled real faces while preserving the identity information during the realism refinement. A lot of photorealistic profile faces are generated and refined by DA-GAN from frontal faces in order to balance the facial pose distribution. The other comes from \cite{yin2017towards}, Face Frontalization Generative Adversarial Network (FF-GAN) focuses on frontalizing faces in the wild under various head poses including extreme profile views. Moreover, a promising method named Disentangled Representation learning Generative Adversarial Network (DR-GAN) from \cite{tran2017representation} endeavors to take the best of both worlds - simultaneously learn pose-invariant identity representation and synthesize faces with arbitrary poses. The recognizers of those models are trained by large dataset, such as FF-GAN has a pre-trained recognizer with CASIA-WebFace, DR-GAN is trained on CASIA-WebFace and AFLW \cite{koestinger2011annotated}. A baseline recognition model of DA-GAN comes from our previous version of TDFF.

\begin{figure}[h]
	\label{Figure6}
	\centering
	\includegraphics[scale=0.52]{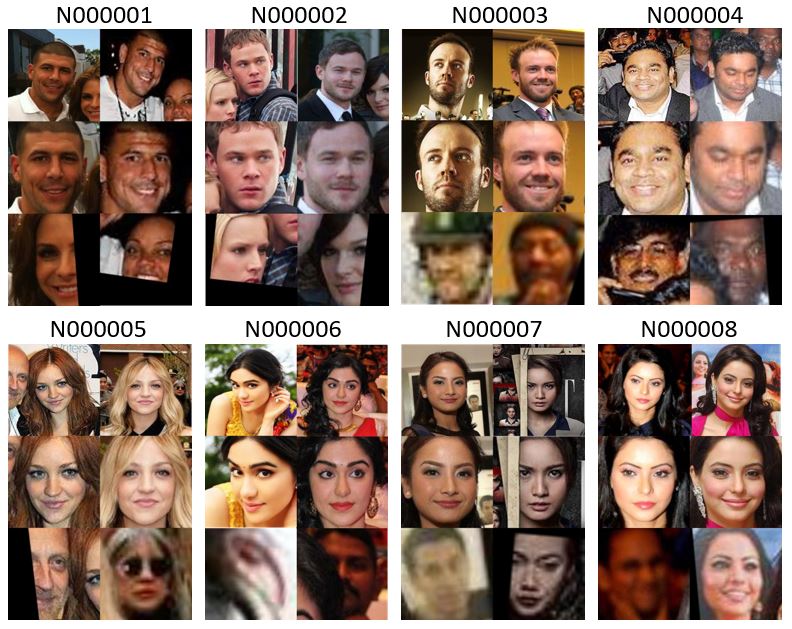}\\
	\caption{{\small Sample face images of our collected and outliers removed. There are eight groups, each of them indicates one subject. The two images of first row are coarsely cropped from collected data, the second row is the refined version of them, the last row represents the filtered outliers.}}
\end{figure}

In the current work, we also follow the similar way--DCNN model should be a good baseline. By virtue of the complementary between different DCNN architectures and datasets, we can obtain a more general feature representation model via ensemble strategy. Intrinsic discrimination of subjects is also important for face recognition, inspired by transfer learning, template specific linear one-vs-rest SVMs are trained in target domain. It shares similar idea as TA \cite{crosswhite2017template} while different negative set is chosen. Similar to \cite{masi2016pose},\cite{hassner2016pooling},\cite{masi2016we}, multiple matching scores are merged into a single one for set-to-set matching whereas an easier way is adopted. Last, we also deploy TPE to further enhance performance of face recognition. More detailed information about our learning framework can be found in the next section part.    

\begin{figure*}[t]
	\label{Figure7}
	\centering
	\includegraphics[scale=0.14]{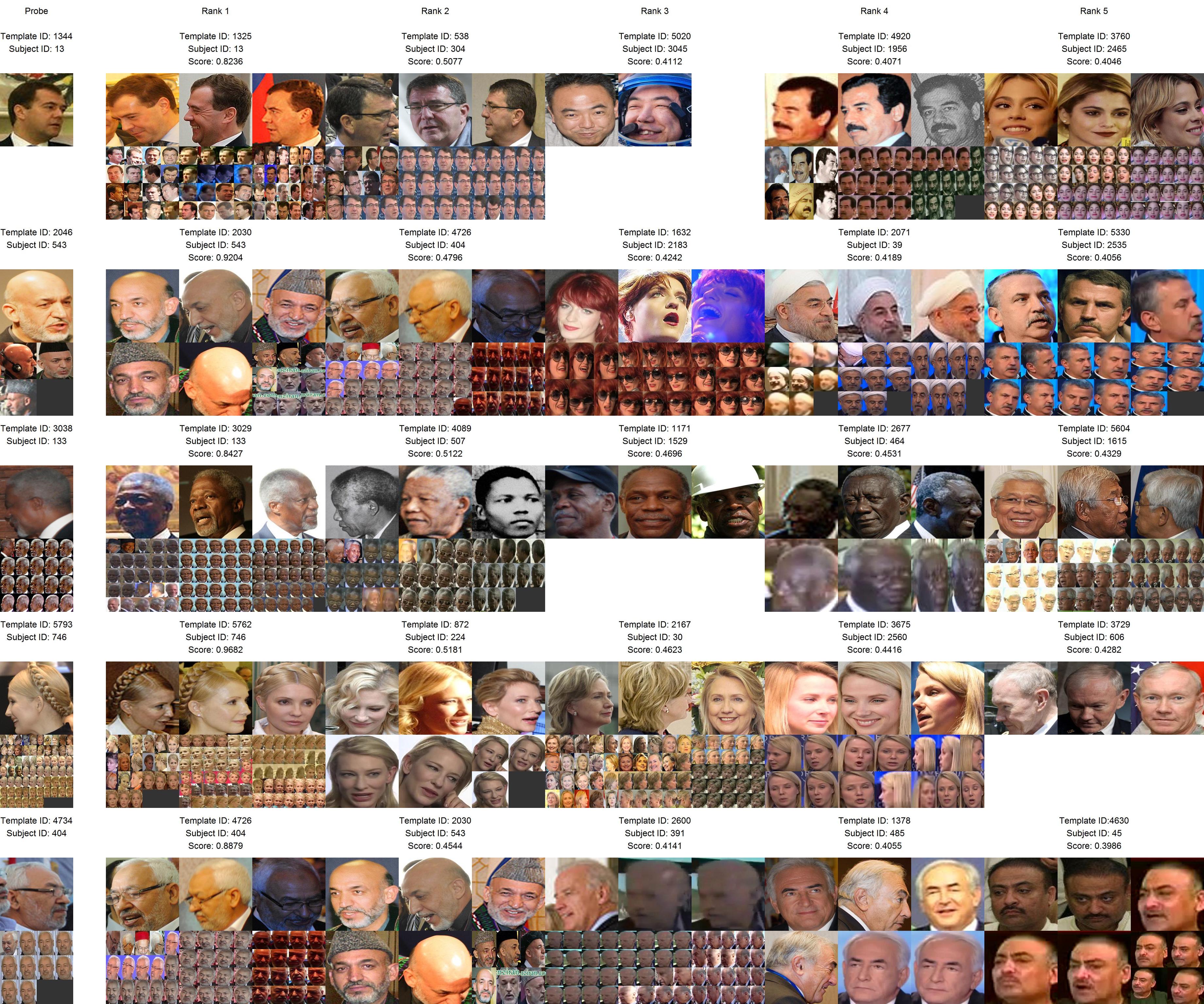}
	\caption{{\small Face identification results for IJB-A split1 on close protocol. The first column shows the query images from probe templates. The remaining 5 columns show the corresponding top-5 queried gallery templates. Subject IDs and Scores are listed on the top of each subject.}}
	\vskip -0.1in
\end{figure*}
\begin{figure*}[t]
	\centering
	\subfloat[The best mated template pairs]
	{\includegraphics[scale=0.055]{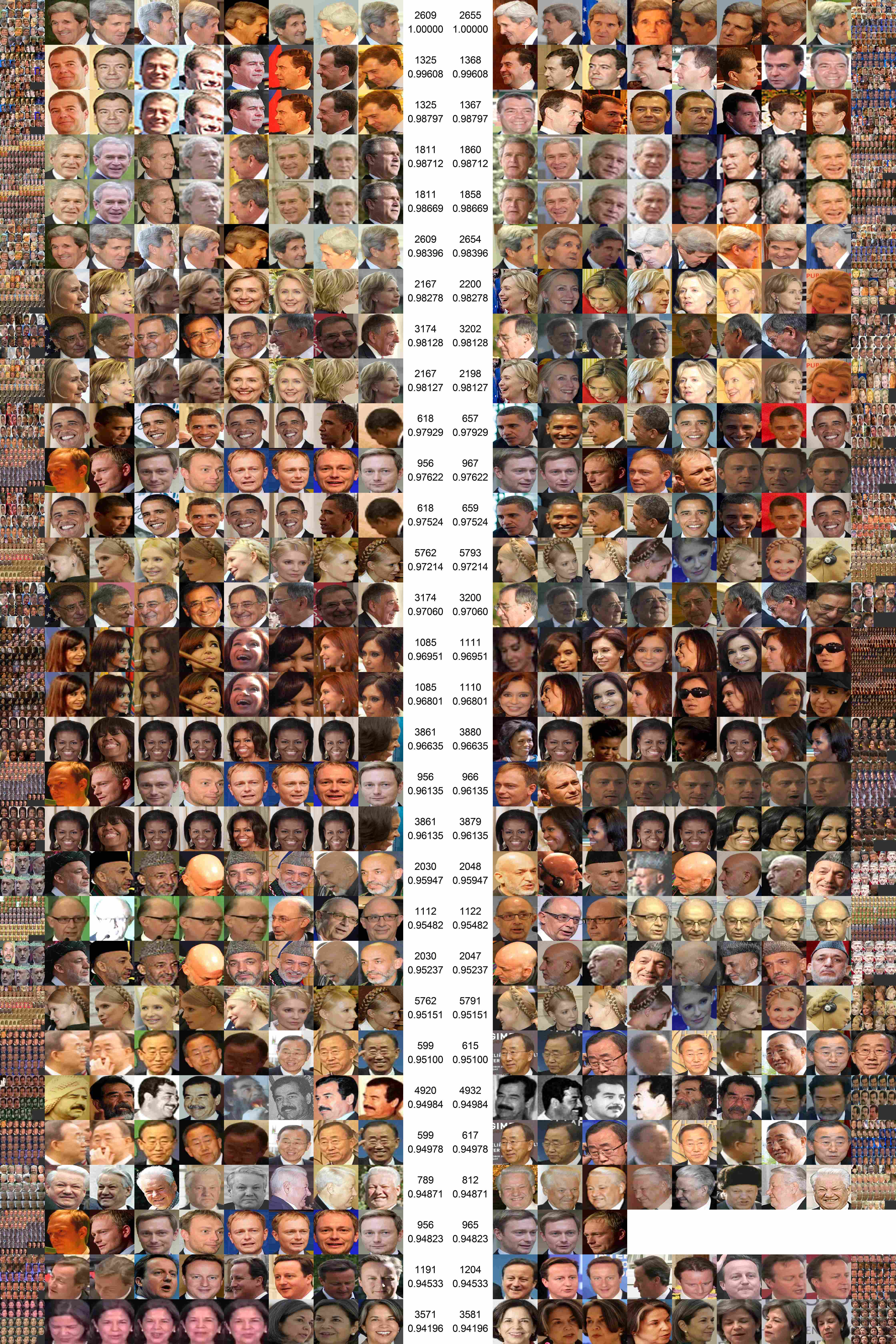}}
	\hspace{0.1in}
	\subfloat[The worst mated template pairs]
	{\includegraphics[scale=0.055]{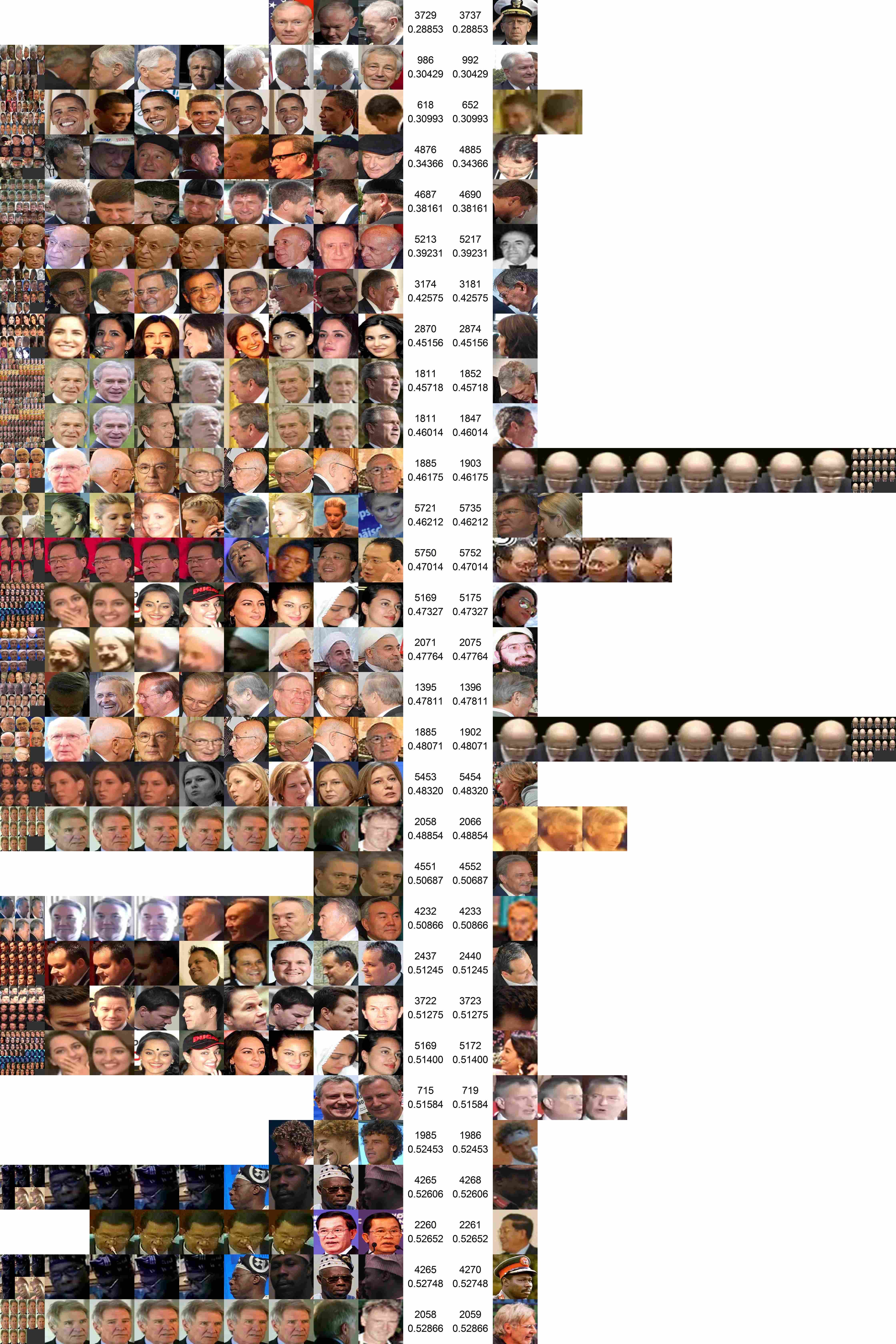}}
	\caption{\label{Figure8} Verification results analysis for mated template pairs on IJB-A split1. In the middle columns of each subfigure, Template IDs and Scores are attached.}
	\vskip -0.1in
\end{figure*}
\section{Transferred Deep Feature Fusion}
\label{sec3:Transferred Deep Feature Fusion}
It is necessary that DCNN architectures are trained on tremendous dataset. However, IJB-A datasets contains 500 subjects with 5,396 images and 2,042 videos sampled to 20,412 frames in total. This is obviously inadequate. Unlike \cite{masi2016we} where training data is increased by synthesizing faces based on pose, 3D shape and expression variations, inspired by domain adaptation, we need other huge labeled face datasets in source domain to train DCNN model. It is different from replacing the final entropy loss layer for a new task and fine-tuning the DCNN model on this new objective using data from the target domain \cite{sharif2014cnn}. We focus on training DCNN model and the one-vs-rest linear SVMs in source domain and target domain, separately. Last, one-shot-similarity (OSS) \cite{wolf2011effective} is utilized to calculate similarity scores and we fuse those multiple matching scores into a single one for final performance evaluation. As shown in Fig.2, our learning framework consists three components: two distinct DCNN models are trained with two different large face datasets in source domain illustrated in middle component, respectively. In target domain, the new unseen data as inputs are fed into those two DCNN architectures with the shared weights and biases learned from source domain for feature extraction, respectively. Then, all features are combined in the first fusion stage. Template specific one-vs-rest SVMs are trained on those fused features in order to boost the intrinsic discrimination of subjects. Last but not least, multiple matching scores computed by OSS is weighted to one final score for verification and identification in the second fusion stage of upper and lower components, respectively. The detailed of each components of our learning framework are presented in the following subsections.   
\vspace{-0.1in}
\subsection{Deep feature learning in source domain}
\label{subsec3.1:Deep feature learning in source domain}
In this part, we discuss detailedly two DCNN models and two extra huge datasets for training in source domain.

Since Network-in-Network (NIN) \cite{lin2013network} has been proposed, the depth of DCNN is refreshed again and again. Recent works \cite{srivastava2015training},\cite{he2016identity},\cite{huang2017densely} have shown that convolutional networks with small filters can be substantially deeper, more accurate, and efficient to train if they contain shorter connections between layers close to the input and those close to the output. The bypassing paths are presumed to be the key factor that eases the training of these very deep networks. This point is further supported by ResNets \cite{he2016deep}, in which pure identity mappings are used as bypassing paths. ResNets have achieved impressive, record-breaking performance on ImageNet \cite{russakovsky2015imagenet}. Until recently, Xie \emph{et al.} \cite{xie2016aggregated} reconstruct the building block of ResNets with aggregating a set of transformations. This simple design results in a homogeneous, multi-branch architecture that has only a few hyper-parameters to set. A new dimension called \emph{cardinality} is proposed, which as an essential factor in addition to the dimension of depth and width. Thus, it is codenamed ResNeXt. A typical block of ResNeXt is shown in Fig.3. Considering the balance between performance and efficiency, we choose ResNeXt 50 (32$\times$4d) as the first DCNN model. 

For public large face dataset, the VGG-Face should be a better choice for ResNeXt 50. The original VGG-Face dataset includes 2,109,307 available images and 2,614 subjects. First, we utilize ground-truth bounding box given by dataset to crop and resize face images from the original ones. Each face image is 144$\times$144. An off-the-shelf CNN model pre-trained on CASIA-WebFace is deployed to do noisy data cleaning. Moreover, the overlap subject with IJB-A dataset should be removed. Finally, we obtain 1,648,187 images and 2,613 subjects in total. For partition of training and validation parts, we refer to ImageNet. 90\% of the total images (1,483,368) are served as training data. 5\% of the total images (82,410) are viewed as validation data. Our implementation for VGG-Face on ResNext 50 is implemented by MXNet \cite{chen2015mxnet}. The image is resized from 144$\times$144 to 480$\times$480 for data augmentation. A 224$\times$224 crop is randomly sampled from 480$\times$480 or its horizontal flip, with the per-pixel mean substracted. The standard color augmentation \cite{krizhevsky2012imagenet} is used. We adopt batch normalization (BN) \cite{ioffe2015batch} right after each convolution and before ReLU. We initialize the weights as in \cite{he2015delving} and train ResNeXt 50 from scratch. NAG with a mini-batch size of 256 is utilized on our GPU cluster machine. The learning rate starts from 0.1 and is divided by 10 every 30 epoch and the model is trained for up to 125 epoch. The weight decay is 0.0001 and the momentum is 0.9. The cardinality is 32. The training and validation curves are shown in Fig.4. Finally, we obtain the validation performance 95.63\% at top1 and 97.00\% at top 5, respectively.  

Inspired by NIN, an orthogonal approach to making networks deeper (e.g., with the help of skip connections) is to increase the network width. The GoogLeNet \cite{szegedy2015going} uses an "Inception module" which concatenates features maps produced by filers of different sizes. Different from ResNext which enhances representational power of network via extremely deep architecture, GoogLeNet depends on wider structure to boost capacity of network. Along with the BN emergence, training DCNN becomes easier than before. Thus, GoogLeNet-BN is our second DCNN model.   

To train GoogLeNet-BN on a much bigger dataset with large number of subjects, data preprocessing is done as following steps. We use OpenCV\cite{pulli2012realtime} to detect face and utilize bounding box to crop and resize face images. Each image is 256$\times$256. There are 582,405 images can not be detected, so we delete them. The overlap subject with IJB-A dataset should be removed. Considering the data distribution, we only keep those identities which have 40-500 images. Finally, we obtain 4,356,052 images and 53,317 subjects in total. Our implementation for our face data on GoogLeNet-BN is implemented by caffe \cite{jia2014caffe}. A 224$\times$224 crop is randomly sampled from 256$\times$256 or its horizontal flip.  We initialize the weights as in \cite{he2015delving} and train GoogLeNet from scratch. SGD with a mini-batch size of 256 is utilized on our GPU cluster machine. The learning rate starts from 0.1 and exp policy is adopted. The weight decay is 0.0001 and the momentum is 0.9. The model are trained for up to 60$\times$$10^{4}$ iterations. We stop training procedure when the error is not decreasing.
\vspace{-0.1in}
\subsection{Template-based unconstrained face recognition in target domain}
\label{subsec3.2:Template-based unconstraied face recognition in target domain}
After finish training procedure of two DCNN models in source domain, weights and biases of ResNext 50 and GoogLeNet-BN are shared into target domain. Each face image or frame of video from target domain is viewed as input to feed into those two models, respectively. For ResNext 50, the penultimate global average pooling layer is served as feature extraction layer. It has 2,048 output size. Thus, the feature dimension is 2,048. Given an image or frame ${{\mathbf{x}}_i} \in {\mathbb{R}^d}$ from a mini-batch of size $M$, where ${d}$ is the dimension of image or frame. ${f_R}\left( {{{\mathbf{x}}_i}} \right) \in {\mathbb{R}^{{d_1}}}$ denotes the feature from ResNeXt 50, where ${d_1} < d$ and ${d_1} = 2048$. Similarly, for GoogLeNet-BN, 7$\times$7 average pooling layer is treated as feature extraction layer. The channel size is 1,024. So, the feature dimension is 1,024. Let ${\text{ }}{f_G}\left( {{{\mathbf{x}}_i}} \right) \in {\mathbb{R}^{{d_2}}}$ is the feature from GoogLeNet-BN, where ${d_2} = 1024$. In the first-stage fusion, ${f_R}\left( {{{\mathbf{x}}_i}} \right)$ and ${f_G}\left( {{{\mathbf{x}}_i}} \right)$ are concatenated into ${\text{ }}{f_F}\left( {{{\mathbf{x}}_i}} \right) \in {\mathbb{R}^{{d_3}}}$, where ${d_3} = 3072$. Finally, each feature is normalized to unit via $L{}_2$ norm for the next procedure.

\begin{table*}[t]\scriptsize
	\caption{\label{Table1} Performance evaluation on the IJB-A dataset. For 1:1 verification, the true accept rates (TAR) @ false positive rates (FAR) are presented. For 1:N identification, the true positive identification rate (TPIR) @ false positive identification rate (FPIR) and CMC are reported}
	\begin{center}
		\begin{tabular}{lccccccccc}
			\toprule
			\multicolumn{1}{c}{\multirow{2}{*}{Method}} &\multicolumn{3}{c}{1:1 Verification TAR} &\multicolumn{1}{c}{} &\multicolumn{5}{c}{1:N Identification TPIR}\\
			\cline{2-4}
			\cline{6-10}
			\multicolumn{1}{c}{}&\multicolumn{1}{c}{FAR=0.001}&\multicolumn{1}{c}{FAR=0.01}&\multicolumn{1}{c}{FAR=0.1} &\multicolumn{1}{c}{} &\multicolumn{1}{c}{FPIR=0.01} &\multicolumn{1}{c}{FPIR=0.1} &\multicolumn{1}{c}{Rank 1} &\multicolumn{1}{c}{Rank 5} &\multicolumn{1}{c}{Rank 10}\\
			\hline
			OpenBR\cite{klontz2013open} & {0.104$\pm$0.014} & {0.236$\pm$0.009} & {0.433$\pm$0.006} & {} &
			{0.066$\pm$0.017} & {0.149$\pm$0.028} & {0.246$\pm$0.011} & {0.375$\pm$0.008} & {-}\\
			GOTS\cite{klare2015pushing} & {0.198$\pm$0.008} & {0.406$\pm$0.014} & {0.627$\pm$0.012} & {} &
			{0.047$\pm$0.024} & {0.235$\pm$0.033} & {0.433$\pm$0.021} & {0.595$\pm$0.020} & {-}\\
			B-CNN\cite{chowdhury2016one} & {-} & {-} & {-} & {} &
			{0.143$\pm$0.027} & {0.341$\pm$0.032} & {0.588$\pm$0.020} & {0.796$\pm$0.017} & {-}\\
			Pooling faces\cite{hassner2016pooling} & {-} & {0.309} & {0.631} & {} & {-} & {-} & {0.846} & {0.933} & {0.951}\\
			LSFS\cite{wang2015face} & 
			{0.514$\pm$0.060} & {0.733$\pm$0.034} &
			{0.895$\pm$0.013} & {} &
			{0.383$\pm$0.063} & {0.613$\pm$0.032} &
			{0.820$\pm$0.024} & {0.929$\pm$0.013} &
			{-}\\
			Deep Multi-pose\cite{abdalmageed2016face} & {-} & {0.787} & {0.911} & {} & {0.52} & {0.75} & {0.846} & {0.927} & {0.947}\\
			DCNN{\tiny{\emph{manual}}}+metric\cite{chen2015end} & {-} & {0.787$\pm$0.043} & {0.947$\pm$0.011} & {} & {-} & {-} & {0.852$\pm$0.018} & {0.937$\pm$0.010} & {0.954$\pm$0.007}\\
			Triplet Similarity\cite{sankaranarayanan2016tripletS} & {0.590$\pm$0.050} & {0.790$\pm$0.030} & {0.945$\pm$0.002} & 
			{} & {0.556$\pm$0.065} & {0.754$\pm$0.014} & {0.880$\pm$0.015} & {0.950$\pm$0.007} & {0.974$\pm$0.006}\\
			VGG-Face\cite{parkhi2015deep} & {-} & {0.805$\pm$0.030} & {-} & {} & {0.461$\pm$0.077} & {0.670$\pm$0.031} & {0.913$\pm$0.011} & {-} & {0.981$\pm$0.005}\\
			PAMs\cite{masi2016pose} & {0.652$\pm$0.037} & {0.826$\pm$0.018} & {-} & {} & {-} & {-} & {0.840$\pm$0.012} & {0.925$\pm$0.008} & {0.946$\pm$0.007}\\
			DCNN{\tiny{\emph{fusion}}}\cite{chen2016unconstrained} & {-} & {0.838$\pm$0.042} & {0.967$\pm$0.009} & {} & {0.577$\pm$0.094} & {0.790$\pm$0.033} & {0.903$\pm$0.012} & {0.965$\pm$0.008} & {0.977$\pm$0.007}\\
			FF-GAN\cite{yin2017towards} & {0.663$\pm$0.033} & {0.852$\pm$0.010} & {-} & {} & {-} & {-} & {0.902$\pm$0.006} & {0.954$\pm$0.005} & {-}\\
			DR-GAN{\tiny{\emph{fuse}}}\cite{tran2017representation} & {0.699$\pm$0.029} & {0.831$\pm$0.017} & {-} & {} & {-} & {-} & {0.901$\pm$0.014} & {0.953$\pm$0.011} & {-}\\
			Masi {\emph{et al.}}\cite{masi2016we} & {0.725} & {0.886} & {-} & {} & {-} & {-} & {0.906} & {0.962} & {0.977}\\
			Triplet Embedding\cite{sankaranarayanan2016tripletP} & {0.813$\pm$0.020} & {0.900$\pm$0.010} & {0.964$\pm$0.005} & {} & {0.753$\pm$0.030} & {0.863$\pm$0.014} & {0.932$\pm$0.010} & {-} & {0.977$\pm$0.005}
			\\
			Template Adaptation\cite{crosswhite2017template} & {0.836$\pm$0.027} & {0.939$\pm$0.013} & {0.979$\pm$0.004} & {} & {0.774$\pm$0.049} & {0.882$\pm$0.016} & {0.928$\pm$0.010} & {0.977$\pm$0.004} & {0.986$\pm$0.003}\\
			Chen {\emph{et al.}}\cite{chen2017unconstrained} & {0.760$\pm$0.038} & {0.889$\pm$0.016} & {0.968$\pm$0.005} & {} & {0.654$\pm$0.001} & {0.836$\pm$0.010} & {0.942$\pm$0.008} & {0.980$\pm$0.005} & {0.988$\pm$0.003}\\
			All-In-One+TPE\cite{ranjan2017all} & {0.823$\pm$0.020} & {0.922$\pm$0.010} & {0.976$\pm$0.004} & {} & {0.792$\pm$0.020} & {0.887$\pm$0.014} & {0.947$\pm$0.008} & {-} & {0.988$\pm$0.003}\\
			NAN\cite{yang2016neural} & {0.881$\pm$0.011} & {0.941$\pm$0.008} & {0.978$\pm$0.003} & {} & {0.817$\pm$0.041} & {0.917$\pm$0.009} & {0.958$\pm$0.005} & {0.980$\pm$0.005} & {0.986$\pm$0.003}\\
			Hayat {\emph{et al.}}\cite{hayat2017joint} & {-} & {-} & {-} & {} & {0.886$\pm$0.041} & {0.960$\pm$0.010} & {0.964$\pm$0.008} & {-} & {1.000$\pm$0.000}\footnotemark\addtocounter{footnote}{-1}\\
			DA-GAN\cite{zhao2017dual} & {0.930$\pm$0.005} & {0.976$\pm$0.007} & {0.991$\pm$0.003} & {} & {0.890$\pm$0.039} & {0.949$\pm$0.009} & {0.971$\pm$0.007} & {0.989$\pm$0.003} & {-}\\
			$L_2$-softmax\cite{ranjan2017l2} & {0.938$\pm$0.008} & {0.968$\pm$0.004} & {0.987$\pm$0.002} & {} & {0.903$\pm$0.046} & {0.955$\pm$0.007} & {0.975$\pm$0.005} & {-} & {0.990$\pm$0.002}\\
			$L_2$-softmax\cite{ranjan2017l2}+TPE\cite{sankaranarayanan2016tripletP} & {0.943$\pm$0.005} & {0.970$\pm$0.004} & {0.984$\pm$0.002} & {} & {0.915$\pm$0.041} & {0.956$\pm$0.006} & {0.973$\pm$0.005} & {-} & {0.988$\pm$0.003}\\
			\hline
			TDFF & {0.919$\pm$0.006} & {0.961$\pm$0.007} & {0.988$\pm$0.003} & {} & {0.878$\pm$0.035} & {0.941$\pm$0.010} & {0.964$\pm$0.006} & {0.988$\pm$0.003} & {0.992$\pm$0.002}\\
			TDFF+TPE\cite{sankaranarayanan2016tripletP} & {0.921$\pm$0.005} & {0.961$\pm$0.007} & {0.989$\pm$0.003} & {} & {0.881$\pm$0.039} & {0.940$\pm$0.009} & {0.964$\pm$0.007} & {0.988$\pm$0.003} & {0.992$\pm$0.003}\\
			TDFF$^*$ & {\bf 0.979$\pm$0.004} & {\bf 0.991$\pm$0.002} & {\bf 0.996$\pm$0.001} & {} & {\bf 0.946$\pm$0.047} & {\bf 0.987$\pm$0.003} & {\bf 0.992$\pm$0.001} & {\bf 0.997$\pm$0.001} & {\bf 0.998$\pm$0.001}\\
			\bottomrule
		\end{tabular}
		\footnotetext{\footnotemark{detailed discussions of the connection between them are shown in our supplementary material.}}
	\end{center}
	\vskip -0.1in
\end{table*}


After feature fusion, in order to train a more discriminative model in target domain, template specific one-vs-rest SVMs play an important role. Specifically, the weights and biases terms for template specific SVMs are learned by optimizing the following ${L_2}$-regularized ${L_2}$-loss objective function:
\begin{equation}
\label{Eqn.1}
	\begin{aligned}
	\mathop {\min }\limits_{\mathbf{w}} \frac{1}{2}{{\mathbf{w}}^T}{\mathbf{w}} + {\lambda _ + }\sum\limits_{i = 1}^{{N_ + }} {\max {{\left[ {0,1 - {y_i}{{\mathbf{w}}^T}{f_F}\left( {{{\mathbf{x}}_i}} \right)} \right]}^2}}  \hfill \\
	+ {\lambda _ - }\sum\limits_{i = 1}^{{N_ - }} {\max {{\left[ {0,1 - {y_j}{{\mathbf{w}}^T}{f_F}\left( {{{\mathbf{x}}_j}} \right)} \right]}^2}}  \hfill \\ 
	\end{aligned}
\end{equation}
where ${\mathbf{w}}$ denote the weights including bias term, ${y_i} \in \left\{ { - 1,1} \right\}$ denotes the label indicating whether the current sample being negative or possible, ${{N_ + }}$ indicates the number of positive samples, ${{N_ - }}$ is the number of negative ones, ${N_ - } \gg {N_ + }$. Moreover, the constraint for negative samples ${\lambda _ - } = C\frac{{{N_ + } + {N_ - }}}{{2{N_ - }}}$, the constraint for positive samples ${\lambda _ + } = C\frac{{{N_ + } + {N_ - }}}{{2{N_ + }}}$, where $C$ is a trade-off factor. A template includes images or/and frames of video. For the feature of video frame, we compute the average media encodings. Let $t_j^V$ denotes average media encoding of video $j$. 
\begin{equation}
\label{Eqn.2}
	t_j^V = \frac{1}{{N_j^V}}\sum\limits_{i = 1}^{N_j^V} {{f_F}\left( {{{\mathbf{x}}_i}} \right)}
\end{equation}
where $N_j^V$ is the number of frame in video $j$,  ${{{\mathbf{x}}_i}}$ denotes $i$ frame of video $j$. In other words, all features of video frames are aggregate one feature. Thus, the deep facial representations for the $a$th template can be expressed as 
\begin{equation}
\label{Eqn.3}
{T_a} = \left\{ {t_i^I,...,t_{{N_a}}^V} \right\}
\end{equation}
where ${t_i^I}$ denotes $i$th image, ${{N_a}}$ express the number of image and video. 
All media encoding need to perform unit normalization. For verification (a.k.a 1:1 compare), the positive sample of template specific SVM is probe template, the large-scale negative samples include the whole training set. For identification (a.k.a 1:N search), the probe template specific SVMs adopt the whole training set as the large-scale negative samples; whereas for gallery template specific SVM, we adopt other gallery templates and the whole training set as large-scale negative samples. Based on One shot similarity (OSS), we compute similarity between two features $p$ and $q$ via $s\left( {p,q} \right) = \frac{1}{2}\mathcal{P}\left( q \right) + \frac{1}{2}\mathcal{Q}\left( p \right)$ where $\mathcal{P}\left( q \right)$ denotes the trained probe template specific SVM model and $\mathcal{Q}\left( p \right)$ indicates the trained gallery template specific SVM model. One template exists many features as Eqn.\ref{Eqn.3}, the resulting multiple matching scores should be ensembled into a single one for each template pair in second-stage fusion.
\begin{equation}
	 s\left( {{T_a},{T_b}} \right) = \frac{{\sum\limits_{{t_i} \in {T_a},{t_j} \in {T_b}} {s\left( {{t_i},{t_j}} \right){e^{\beta {\text{ }}s\left( {{t_i},{t_j}} \right)}}} }}{{\sum\limits_{{t_i} \in {T_a},{t_j} \in {T_b}} {{e^{\beta {\text{ }}s\left( {{t_i},{t_j}} \right)}}} }}
\end{equation}
where $\beta  = 0$ is enough in our following experiments.

\subsection{New features based on new data and advanced neural architecture}
\label{subsec3.3:New features based on new data and advanced neural architecture}
Recently, Hu {\emph{et al.}} \cite{hu2017squeeze} proposes the Squeeze-and-Excitation (SE) block, which adaptively recalibrates channel-wise feature responses by explicitly modelling interdependencies between channels. Specifically, the basic structure of the SE building block can be constructed to perform feature recalibration as follows. The feature maps are first passed through a squeeze operation, which aggregates the feature maps across spatial dimensions to produce one channel descriptor. it enables information from the global receptive field of the network to be utilized by its following layers. Then it is followed by an excitation operation where a self-gating mechanism is deployed to learn channel dependency. Last, the feature maps are
reweighed to generate the output of the SE building block and then it can be fed directly into the subsequent layers. This procedure is depicted as blue box in Fig.5. We integrate SE building block to ResNext block as illustrated in Fig.5. Finally, SE-ResNeXt 101 (64$\times$4d) is deployed in our framework as other DCNN model. 

In order to train the very deep neural network of SE-ResNeXt 101 and cater to the similar setting of IJB-A such as large pose variations, we collect new large face dataset via Google Image Search and detect them by the model of \cite{zhang2016jointface}. After preprocessing by multiple detectors such as OpenCV \cite{pulli2012realtime} and MTCNN \cite{zhang2016jointface} and cleaning outliers by our pre-trained ResNeXt 101 model trained on our previously collected large dataset, we obtain around 10000 subjects and $\mathcal{O}\left( 10^6 \right)$ images in total. In Fig.6, we illustrate some sample images of this new large face dataset and some outliers removed by our pre-trained model with proper threshold. During training progress, we deploy more data augmentation skills such as random contrast, brightness and saturation in order to fit the large illumination variation of IJB-A as much as possible. Before training SE-ResNeXt 101, we remove the overlapping subjects with IJB-A first, then normalize and rescale input image to 122$\times$144, then resize them to 256 on short one between height and width with keeping aspect ratio for data augmentation. Other settings are the same as training ResNeXt 50 on VGG-Face, except the mini-batch of 128 is applied on our DGX-1 with 8 GPUs.



\begin{table}[h] 
	\caption{\label{Table2} Performance evaluation on the IJB-A dataset. For 1:1 verification, the true accept rates (TAR) @ false positive rates (FAR) are presented. }
	\begin{center}
		\begin{tabular}{lc}
			\toprule
			\multicolumn{1}{c}{\multirow{2}{*}{Method}} &\multicolumn{1}{c}{1:1 Verification TAR}\\
			\cline{2-2}
			\multicolumn{1}{c}{}&\multicolumn{1}{c}{FAR=0.0001}\\
			\hline
			$L_2$-softmax(FR)\cite{ranjan2017l2} & {0.832$\pm$0.027} \\
			$L_2$-softmax(FR)\cite{ranjan2017l2}+TPE\cite{sankaranarayanan2016tripletP} & {0.863$\pm$0.012} \\
			$L_2$-softmax(R101)\cite{ranjan2017l2} & {0.879$\pm$0.028} \\
			$L_2$-softmax(R101)\cite{ranjan2017l2}+TPE\cite{sankaranarayanan2016tripletP} & {0.898$\pm$0.019}\\
			$L_2$-softmax(RX101)\cite{ranjan2017l2} & {0.883$\pm$0.032}\\
			$L_2$-softmax(RX101)\cite{ranjan2017l2}+TPE\cite{sankaranarayanan2016tripletP} & {0.909$\pm$0.007}\\
			\hline
			TDFF & {0.875$\pm$0.013} \\
			TDFF+TPE\cite{sankaranarayanan2016tripletP} & {0.877$\pm$0.018} \\
			TDFF$^*$ & {\bf 0.959$\pm$0.014} \\
			\bottomrule
		\end{tabular}
	\end{center}
	\vskip -0.1in
\end{table}

\begin{figure*}[h]
	\centering
	\subfloat[The best nonmated template pairs]
	{\includegraphics[scale=0.055]{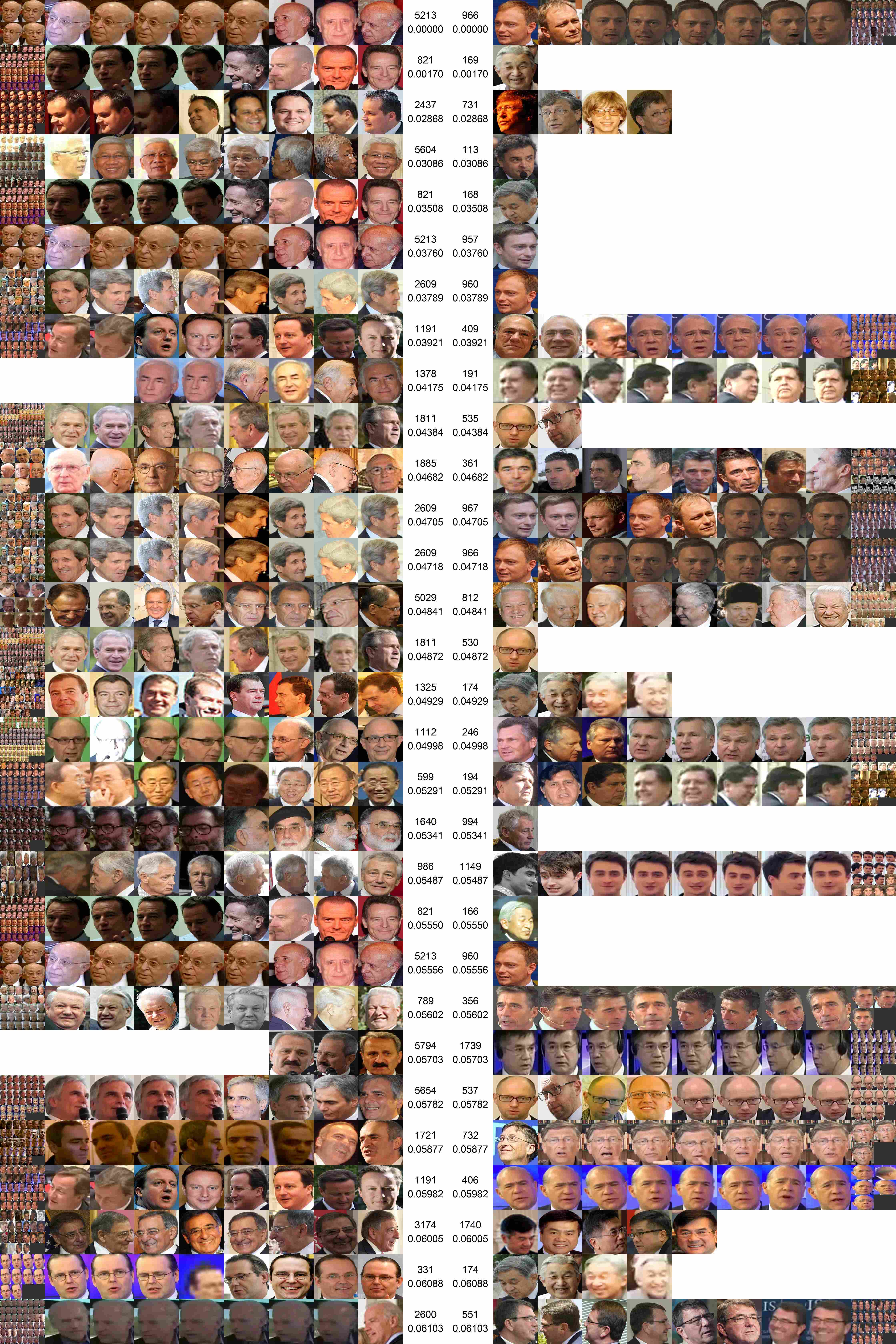}}
	\hspace{0.1in}
	\subfloat[The worst nonmated template pairs]
	{\includegraphics[scale=0.055]{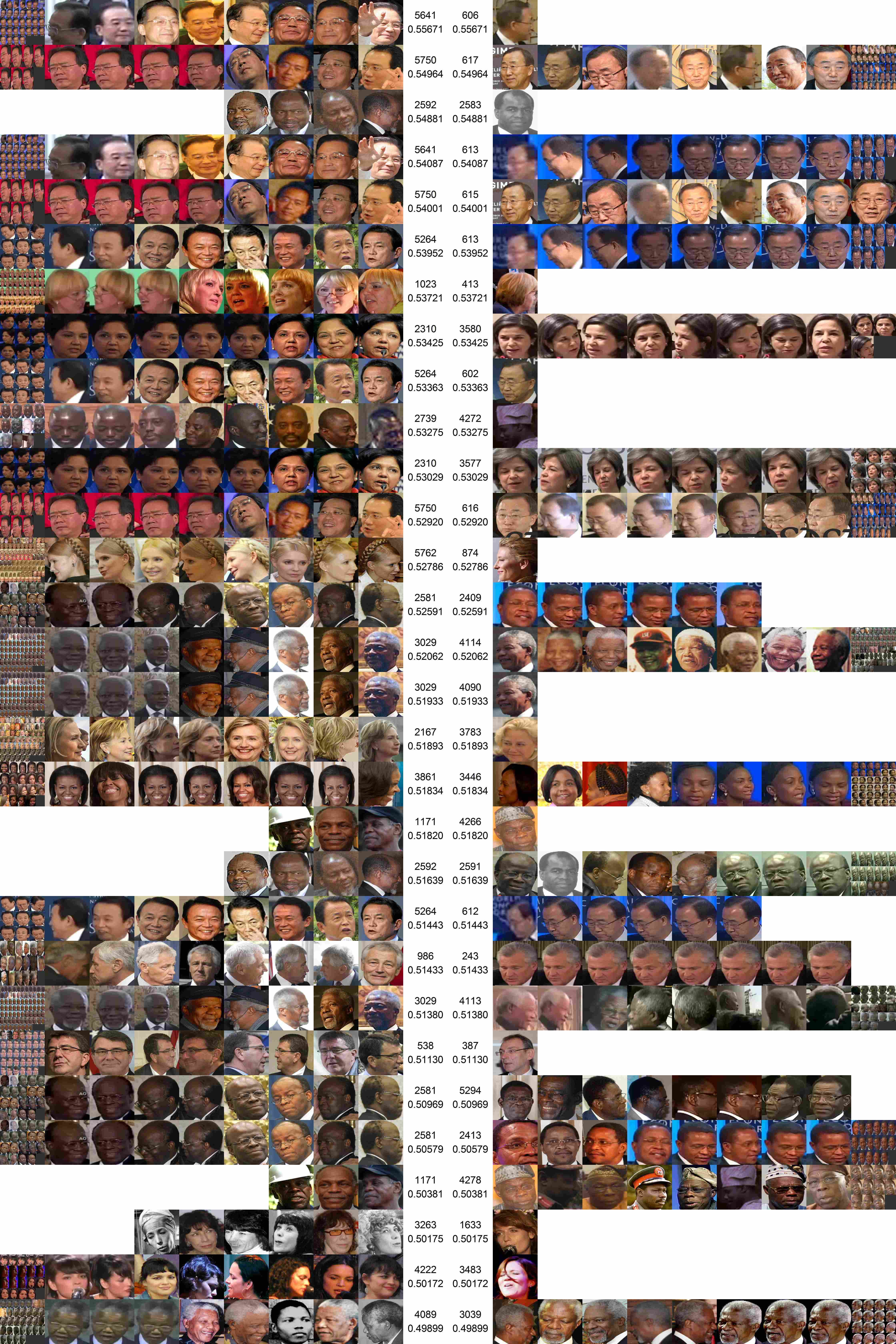}}
	\caption{\label{Figure9} Verification results analysis for nonmated template pairs on IJB-A split1. In the middle columns of each subfigure, Template IDs and Scores are attached.}
	\vskip -0.1in
\end{figure*}

\begin{figure*}[h]
	\centering
	\subfloat[The worst nonmated template pairs from TDFF]
	{\includegraphics[scale=0.055]{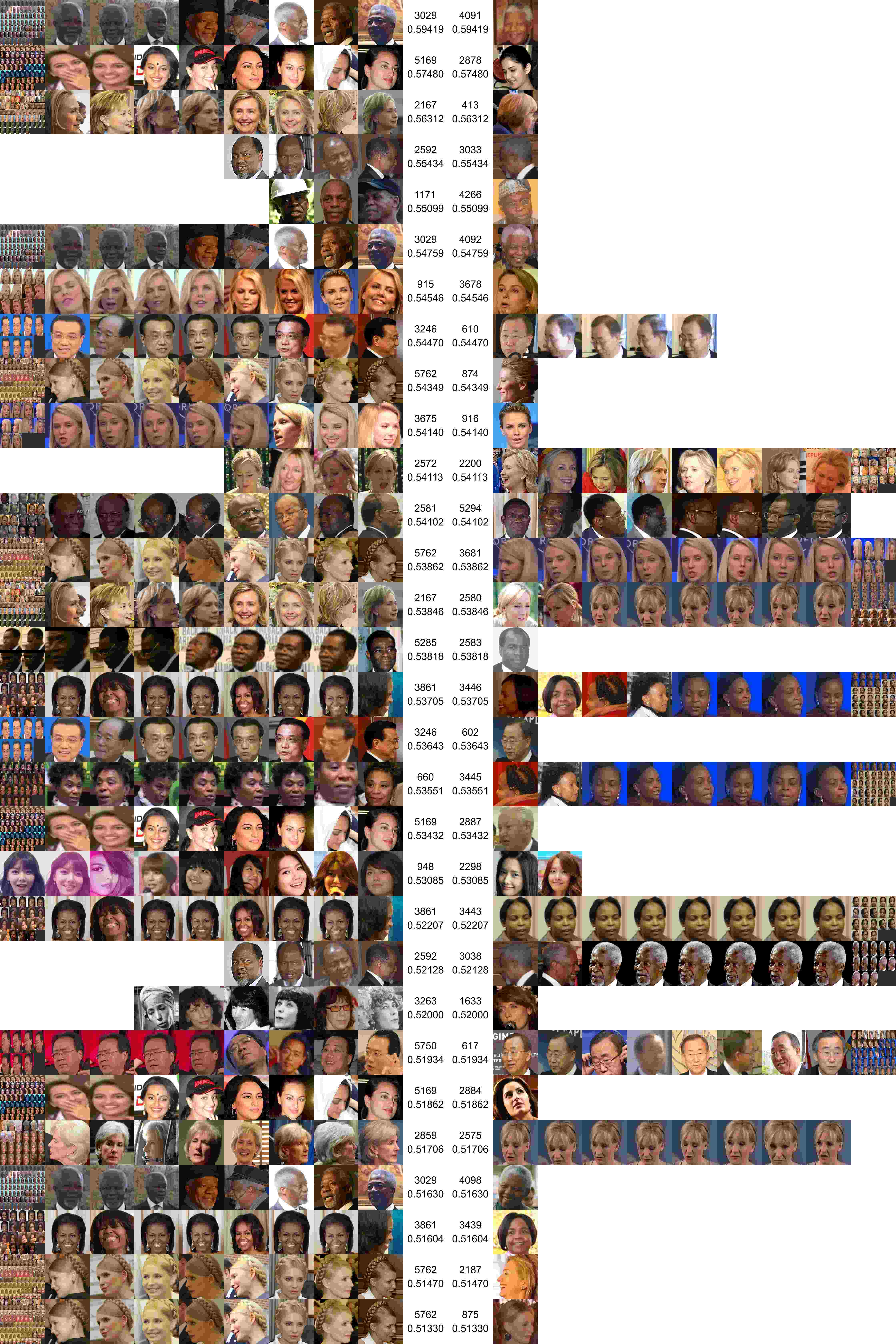}}
	\hspace{0.1in}
	\subfloat[The worst nonmated template pairs from TDFF$^*$]
	{\includegraphics[scale=0.055]{worst_nonmated_SE_RSX101_RSX152_v2.jpg}}
	\caption{\label{Figure10} Comparison between TDFF and TDFF$^*$ on results of worst nonmated template pairs of IJB-A split1 for verification. All scores of TDFF$^*$ are lower than that of TDFF, the lower the better for worst nonmated setting.}
	\vskip -0.1in
\end{figure*}

\section{Experiments and analysis}
\label{sec4:Experiments and analysis}
In this section, we describe the results for evaluation of the experimental system on the IJB-A verification and identification protocols. The IJB-A dataset contains face images and video frames captured from unconstrained settings which are aligned better with the requirements of real applications. There are 500 subjects with 5,396 images and 2,042 videos sampled to 20,412 frames in total. Full pose variation and wide variations in imaging conditions are the main features of IJB-A dataset, which makes the face recognition very challenging. In our experiments, we just utilize the ground-truth bounding box to crop face image from the original one and resize to 224$\times$224 for each image or frame. We do not use any off-the-shelf pre-trained DCNN model to clean data. We also do not deploy any face detector and do not perform any face alignment procedure. 

A remarkable feature of this dataset is that the concept of template is introduced. Each training or testing sample is called a template which comprises a mixture of static images and sampled video frames. Each static image or a frame of video corresponds with a media. On average, each subject has 11.4 images and 4.2 videos. There are 10 training and testing splits. Each of them contains 333 and 167 subjects, respectively. 

In Table \ref{Table1}, we list the performance of state-of-the-art algorithms on IJB-A dataset, where $^1$ denotes the author may not utilize the ground-truth bounding box of IJB-A dataset, because we find there are some errors or noises in that. When we use the TPE to learn a discriminative mapping space while keep the original feature dimension using the training splits of IJB-A. It slightly improves the performance and achieves the better one with TAR of 0.921 @ FAR = 0.001, TAR of 0.961 @ FAR = 0.01 and TAR of 0.989 @ FAR = 0.1 for verification. 

Last but not least, we fuse two new features from not only SE-ResNeXt 101 and ResNeXt 152 trained on our newly collected large face datasets. Our performance denoted with $*$ achieves the best of them for both verification and identification protocols with large gap. Specifically, we obtain the best performance with TAR of 0.979 @ FAR = 0.001, TAR of 0.991 @ FAR = 0.01 and TAR of 0.996 @ FAR = 0.1 for verification and TPIR of 0.946 @ FPIR = 0.01, TPIR 0f 0.987 @ FPIR = 0.1 for identification open protocol. Based on our new training data, advanced neural architecture and more reasonable data augmentation skills, our framework performs significantly more even better than state-of-the-art algorithms in all protocols. These results clearly suggest the effectiveness of our proposed learning framework. In \cite{ranjan2017l2}, the author reports the results for a very low FAR at 0.0001. Thus, in Table \ref{Table2}, we also report the performance @ FAR = 0.0001 for verification protocol, our results still the best than $L_2$-softmax, even without TPE.

We illustrate the identification results for IJB-A split1 on close protocol in Fig.7. The first column shows the query images from probe templates. The remaining 5 columns show the corresponding top-5 queried gallery templates. For each template, we provide Template ID, Subject ID and similarity score. For all five rows, our approach can successfully find the subjects in rank 1. 

Finally, we visualize the verification results in Fig.8 and Fig.9 for IJB-A split1 to gain insight into template based unconstrained face recognition. After computing the similarities for all pairs of probe and reference templates, we sort the resulting list. Each row represents a probe and reference template pair. The original templates within IJB-A contain from one to dozens of media. Up to eight individual media are shown with the last space showing a mosaic of the remaining media in the template. Between the templates are the Template IDs for probe and reference as well as the best mated and best non-mated similarity. Fig.8 (a) shows the highest mated similarities. In the thirty highest scoring correct matches, we note that every reference template contains dozens of media. The probe templates also contain dozens of media that matches well. Fig.8 (b) shows the lowest mated template pairs, representing failed matching. The thirty lowest mated results from single-media reference templates are under extremely challenging unconstrained conditions. There extremely difficult cases cannot be solved even using our proposed approach. Fig.9 (a) showing the best non-mated similarities shows the most certain nonmates, again often involving large templates with enough guidance from the relevant and historical information. Fig.9 (b) showing the worst non-mated pairs highlights the unstable errors involving single-media reference templates representing impostors in challenging orientation. Last, we illustrate the comparison between TDFF and TDFF$^*$ on results of worst nonmated template pairs of IJB-A split1 for verification in Fig.10. The scores should the lower the better. From this view, it also demonstrates the performance of TDFF$^*$ is better than that of TDFF.

\section{Conclusion}
\label{sec5:Conclusion}
In this paper, we propose a unified learning framework named transferred deep feature fusion. It can effectively integrate superiority of each module and outperform the state-of-the-art on IJB-A dataset. Inspired by transfer learning, facial feature encoding model of subjects are trained offline in a source domain, and this feature encoding model is transferred to a specific target domain where limited available faces of new subjects can be encoded. Specifically, in order to capture the intrinsic discrimination of subjects and enhance the generalization capability of face recognition models, we deploy two advanced deep convolutional neural networks (DCNN) with distinct architectures to learn the representation of faces on two different large datasets (each one has no overlap with IJB-A dataset) in source domain. These two DCNN models provide distinct feature representations which can better characterize the data distribution from different perspectives. The complementary between two distinct models is beneficial for feature representation. Thus, representing a face from different perspectives could effectively decrease ambiguity among subjects and enhance the generalization performance of face recognition especially on extremely large number of subjects. After offline training procedure, those two DCNN models are transferred to target domain where templates of IJB-A dataset as inputs are performed feature extraction with shared weights and biases, respectively. Then, two-stage fusion is designed. Features from two DCNN models are combined in order to obtain more discriminative representation in first-stage. Then, template specific linear SVMs are trained on fused features for classification. Finally, for set-to-set matching problem, multiple matching scores are merged into a single one for each template pair as the final results in the second-stage of fusion. Comprehensive evaluations on IJB-A public dataset well demonstrate the significant superiority of the proposed learning framework. Based on the proposed approach, we have submitted our IJB-A results to NIST for official evaluation. Furthermore, by introducing new data and advanced neural architecture, our method outperforms the state-of-the-art by a wide margin on IJB-A dataset. In the future, end-to-end network architecture is still attractive for face recognition. Manifold-based metric learning can learn non-linear embedding space, it can explore the geometric structure of the feature encoding. Because, the rotation of head follows a low-dimension manifold. Dictionary learning combines DCNN is an interesting task.



\begin{IEEEbiography}
[{\includegraphics[width=1in,height=1.25in,clip,keepaspectratio]{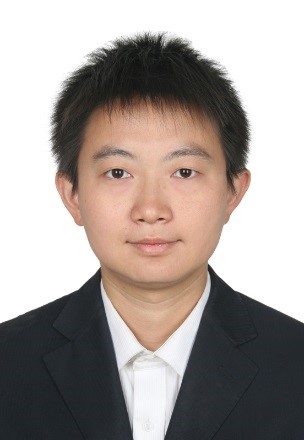}}]
	{Lin Xiong}
received the B.S. degree from Shaanxi University of Science \& Technology in 2003, and he received the Ph.D. degree with School of Electronic Engineering, Xidian University, China, in 2014. He is currently a research engineer of Learning \& Vision, Core Technology Group, Panasonic R\&D Center Singapore, Singapore. His current research interests include unconstrained/large-scale face recognition, person re-identification, deep learning architecture engineering, transfer learning, Riemannian manifold optimization, sparse and low-rank matrix factorization.
\end{IEEEbiography}

\begin{IEEEbiographynophoto}
	{Jayashree~Karlekar} 	
\end{IEEEbiographynophoto}

\begin{IEEEbiography}
	[{\includegraphics[width=1in,height=1.25in,clip,keepaspectratio]{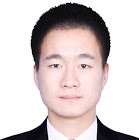}}]
	{Jian~Zhao} received the B.S. degree from Beihang University in 2012, and he received the Master degree with School of Computer, National University of Defense Technology, China, in 2014. He is currently funded by China Scholarship Council (CSC) and School of Computer, National University of Defense Technology to pursue his Ph.D. degree at Learning and Vision Group, Department of Electronical and Computer Engineering, Faculty of Engineering, National University of Singapore. His current research interests include face recognition, human parsing, human pose estimation, object detection, object semantic segmentation, and relevant deep learning and computer vision problems.
\end{IEEEbiography}

\begin{IEEEbiography}
	[{\includegraphics[width=1in,height=1.25in,clip,keepaspectratio]{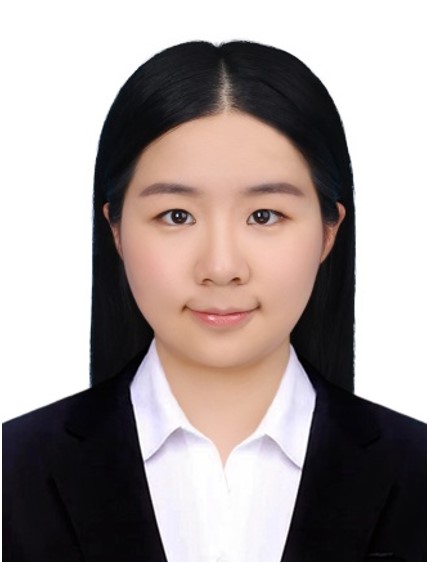}}]
	{Yi~Cheng} received the B.S. degree from Wuhan University in 2016 and the Master degree from National University of Singapore in 2017. She is currently a research engineer of Learning \& Vision, Core Technology Group, Panasonic R\&D Center Singapore, Singapore. Her research is focused on implementing deep learning algorithms on object detection and face recognition.
\end{IEEEbiography}

\begin{IEEEbiography}
	[{\includegraphics[width=1in,height=1.25in,clip,keepaspectratio]{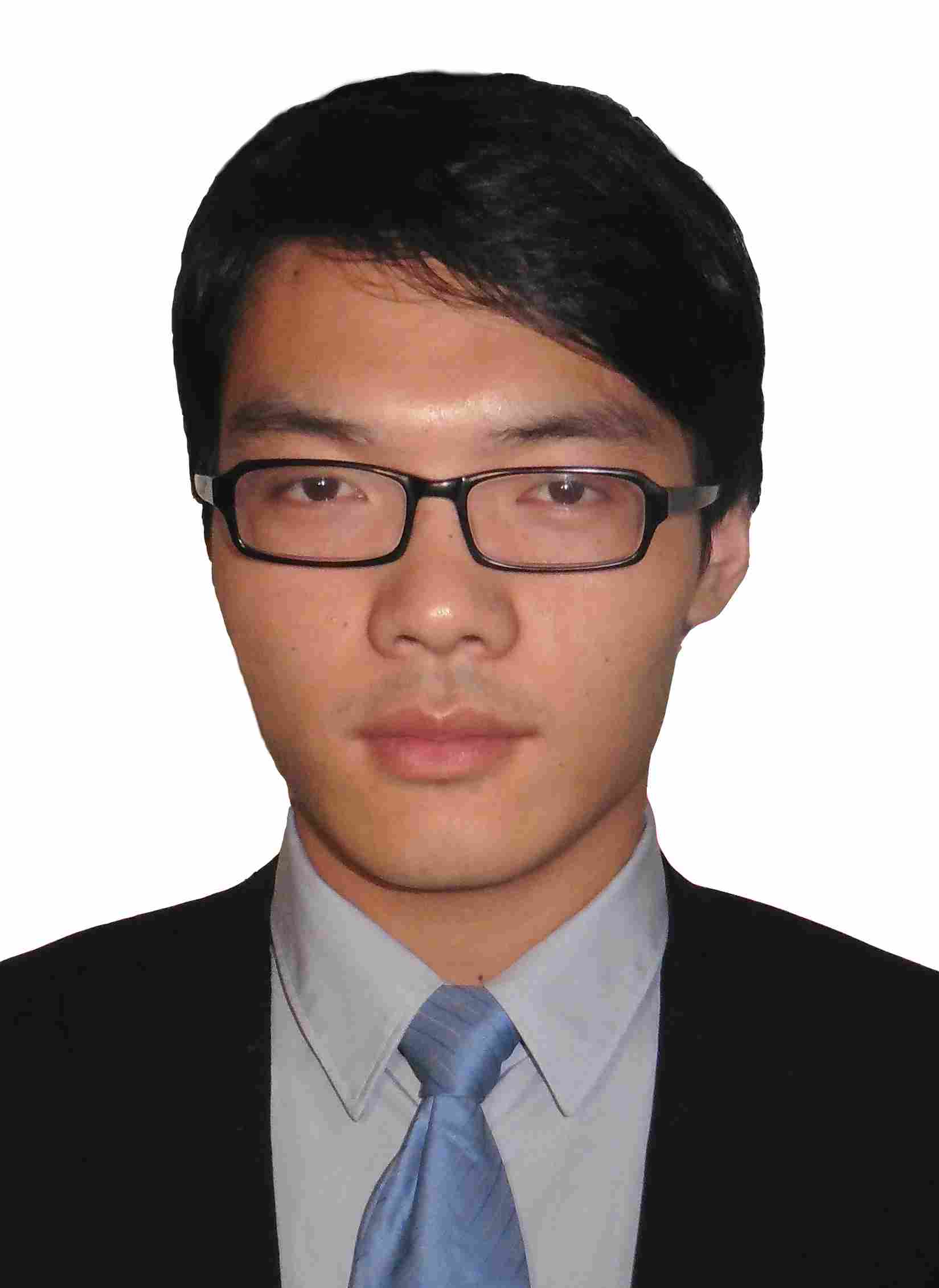}}]
	{Yan~Xu} received the B.S. degree from Lanzhou University of Technology in 2012, and the Masters degree from Xidian University in 2015. He is currently a research engineer of Learning \& Vision, Core Technology Group,  Panasonic R\&D Center Singapore, Singapore. His research interests include unconstrained/large-scale/low-shot face verification/identification, facial landmark localization, and deep learning architecture engineering.
\end{IEEEbiography}

\begin{IEEEbiography}
	[{\includegraphics[width=1in,height=1.25in,clip,keepaspectratio]{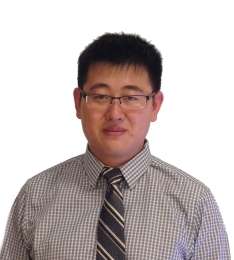}}]
	{Jiashi~Feng} is currently an assistant Professor in the department of electrical and computer engineering in the National University of Singapore. He got his B.E. degree from University of Science and Technology, China in 2007 and Ph.D. degree from National University of Singapore in 2014. He was a postdoc researcher at University of California from 2014 to 2015. His current research interest focus on machine learning and computer vision techniques for large-scale data analysis. Specifically, he has done work in object recognition, deep learning, machine learning, highdimensional statistics and big data analysis.
	
\end{IEEEbiography}

\begin{IEEEbiographynophoto}
	{Sugiri~ Pranata}	
\end{IEEEbiographynophoto}

\begin{IEEEbiographynophoto}
	{Shengmei~ Shen}	
\end{IEEEbiographynophoto}

\end{document}